\documentclass[runningheads]{llncs}
\usepackage{graphicx}
\usepackage{amsmath,amssymb}
\usepackage{color}
\usepackage{overpic}
\usepackage{amssymb}
\usepackage{graphicx}
\usepackage{mathrsfs}
\usepackage{xcolor}
\usepackage{float} %< to define algorithms
\usepackage{anyfontsize} %< remove warnings
\usepackage{sidecap} %< caption on side
\usepackage{microtype} 
\usepackage{booktabs} 
\usepackage{diagbox}

\definecolor{turquoise}{cmyk}{0.65,0,0.1,0.3}
\definecolor{purple}{rgb}{0.65,0,0.65}
\definecolor{dark_green}{rgb}{0, 0.5, 0}
\definecolor{orange}{rgb}{0.8, 0.6, 0.2}
\definecolor{red}{rgb}{0.8, 0.2, 0.2}
\definecolor{dark_blue}{rgb}{0.0, 0.1, .7}
\definecolor{light_blue}{rgb}{0.3, 0.3, .9}
\definecolor{dark_cyan}{cmyk}{1.0, 0.0, 0.0, 0.1}
\definecolor{light_gray}{rgb}{0.7, 0.7, .7}
\definecolor{pink}{rgb}{1, 0, 1}
\definecolor{accent}{rgb}{179,81,109}
\definecolor{anagreen}{rgb}{.13,.627,.494}
\definecolor{anasalmon}{rgb}{.85,.604,.564}

\newcommand{\hidden}[1]{{}} %< discards

\newcommand{\ADDRESSED}[1]{{}}

\definecolor{darkolivegreen}{rgb}{0.5, 0.7, 0.3}

\newcommand{\Figure}[1]{Figure~\ref{fig:#1}}

\newcommand{\Sec}[1]{Sec.~\ref{sec:#1}}
\newcommand{\Section}[1]{Section~\ref{sec:#1}}

\newcommand{\brief}[1]{} % ONLY LATEX VISIBLE
\usepackage{currfile}

\newcommand{\myfigurename}{\put(-4,0){\large \vertical{\textbf{\todo{\currfiledir}}}}}
\renewcommand{\myfigurename}{}

\floatstyle{boxed}
\newfloat{algorithm}{t}{alg}
\floatname{algorithm}{Table}

\newcommand{\Table}[1]{Table~\ref{tab:#1}}

\renewcommand{\paragraph}[1]{\vspace{.04in} \noindent \textbf{#1.}}

\newcommand{\citet}{\cite}

\usepackage{snapshot}

\begin{document}
\pagestyle{headings}
\mainmatter

%\def\ACCV18SubNumber{641}
%===========================================================
\title{HandSeg: An Automatically Labeled Dataset for Hand Segmentation from Depth Images}
%\titlerunning{ACCV-18 submission ID \ACCV18SubNumber}
%\authorrunning{ACCV-18 submission ID \ACCV18SubNumber}
%\author{Anonymous ACCV 2018 submission}
%\institute{Paper ID \ACCV18SubNumber}
\author{
Abhishake Kumar Bojja$^\ast$
\and
Franziska Mueller$^\dagger$
\and
Sri Raghu Malireddi$^\ast$
\and
\newline
Markus Oberweger$^\ddagger$
\and
Vincent Lepetit$^\ddagger$
\and
Christian Theobalt$^\dagger$
\and
\newline
Kwang Moo Yi$^\ast$
\and
Andrea Tagliasacchi$^\ast$
\newline
%$\ast$-University of Victoria, $\dagger$-MPI Informatics, $\ddagger$-TU Graz\\
{\tt\small \{abojja, raghu, kyi, ataiya\}@uvic.ca}\\
{\tt\small \{frmueller, theobalt\}@mpi-inf.mpg.de}\\
{\tt\small \{oberweger, lepetit\}@icg.tugraz.at}
}
\institute{$\ast$-University of Victoria, $\dagger$-MPI Informatics, $\ddagger$-TU Graz}

\titlerunning{HandSeg: Dataset for Hand Segmentation from Depth Images}
%: An Automatically Labeled Dataset for Hand Segmentation from Depth Images}
\authorrunning{Bojja, Mueller, Malireddi, Oberweger, Lepetit, Theobalt, Yi, Tagliasacchi}

\maketitle
\begin{abstract}
We propose an automatic method for generating high-quality annotations for depth-based hand segmentation, and introduce a large-scale hand segmentation dataset.
Existing datasets are typically limited to a single hand.
By exploiting the visual cues given by an RGBD sensor and a pair of colored gloves, we automatically generate dense annotations for two hand segmentation. 
This lowers the cost/complexity of creating high quality datasets, and makes it easy to expand the dataset in the future.
We further show that existing datasets, even with data augmentation, are not sufficient to train a hand segmentation algorithm that can distinguish two hands.
Source and datasets will be made publicly available.
\end{abstract}
%%% Local Variables:
%%% mode: latex
%%% TeX-master: "../paper"
%%% End:

\begin{figure}[t]
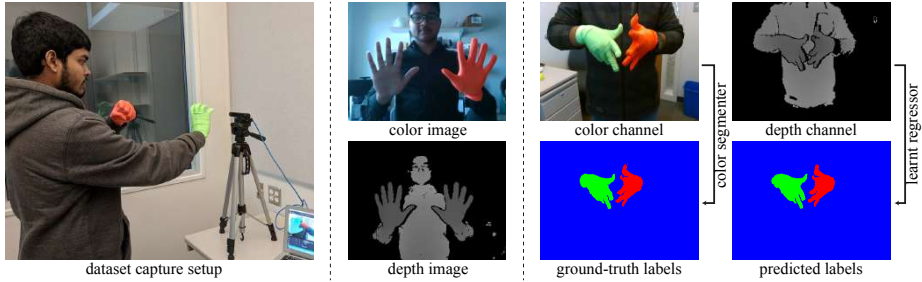

\centering
\begin{overpic} 
[width=\linewidth]
% [width=\linewidth,grid,tics=10]
{\currfiledir/item-1.pdf}
\myfigurename{}
\end{overpic}
\caption{
Proposed data capture and automatic annotation framework. 
{\bf(Left)}~Our dataset is constructed by recording a user performing hand movements wearing a pair of brightly colored gloves in front of a depth camera.
To the best of our knowledge, our dataset is the first \emph{two-hand} dataset for hand segmentation.
{\bf(Middle)}~The use of tight colored gloves provide a \emph{quasi} non-invasive automatic annotation system, as the signal-to-noise ratio of a conventional depth sensor is not sufficiently high to distinguish between gloved and bare hands.
{\bf(Right)}~Color images that are aligned with the depth images are exploited to automatically compute ground truth labels without user intervention.
We then quickly filter out the few wrongly labeled images through human inspection. We can subsequently use these input-label pairs to train a depth-based semantic segmenter.
}
\label{fig:teaser}
\end{figure}
%%% Local Variables:
%%% mode: latex
%%% TeX-master: "../../paper"
%%% End:

\section{Introduction}
Hand gestures are a natural way for humans to interact with the surrounding environment, and as such, many researchers have focused on obtaining accurate hand poses~\cite{erol2007vision,supancic2015depth}. Recently, as depth cameras have become more accurate and affordable~\cite{keselman2017intel,fanello17}, substantial progress has been made towards this goal~\cite{tagliasacchi2015robust,oberweger2015hands,sridhar2015fast}. In many cases, the first step in obtaining accurate poses of hands is to find \emph{where} the hand is in the image, preferably as accurately and robustly as possible.
In \emph{hand segmentation}, the detection happens at pixel-level accuracy.

A number of heuristic solutions have been proposed to simplify the task of hand
segmentation~\cite{oberweger2015hands,tagliasacchi2015robust,sharp2015accurate}. While these approaches are well suited for small-scale lab experiments, they do not possess the robustness required for a consumer-level solution that needs to work under the full diversity of interactions in general real-world scenes -- the violation of one of their underlying assumptions results in immediate tracking failure. 
One could learn a hand segmenter from a dataset of annotated depth images. 
However, as we will show, the limited size and quality of currently available datasets results in segmenters that typically overfit to the training data, and do not generalize well to unseen scenarios. Due to the limited size of available datasets, the application of modern deep learning solutions to the problem of real-time hand segmentation has received limited attention.

Hence, a central goal is to capture a sufficiently \emph{large} dataset equipped with \emph{high-quality} ground truth annotations.
To achieve this, we propose an automatic procedure to create high-quality per-pixel hand segmentation annotations from depth data, and introduce a large-scale dataset that we captured and annotated using the proposed method.
As shown in \Figure{teaser}, we obtain this dataset by having a number of users perform hand gestures in front of an RGBD camera while wearing a pair of \emph{colored gloves}. The color and depth channels are then used to generate high-quality ground truth annotations with minimal user intervention.

Note that the only additional equipment necessary for data acquisition is a pair of colored gloves, compared to the sophisticated setups used for hand capture (magnetic sensors~\cite{yuan2017bighand} or optical IR markers~\cite{opticalhand}).
Moreover, the quality of the dataset is much better than the ones that use motion capture sensors, as these methods require an additional heuristics to generate pixel-wise annotations for training a hand segmenter~\cite{wetzler2015rule}.
To the best of our knowledge, our dataset is the only one that provides both quality and quantity, with the quantity being orders of magnitude larger than what is currently available~(see \Table{datasets}). We also provide an in-depth analysis of the effect of using our dataset on multiple neural network architectures for hand segmentation, as well as traditional Random Forests due to their computational efficiency.
We empirically find that using \emph{strided \mbox{[transposed-]}convolutions} in place of [un]pooling layers, and the use of skip-connections is essential for achieving high-accuracy.
This further enables efficient forward-passes within $\approx 5ms$ on an NVIDIA Geforce GTX1080 Ti, making our approach suitable for real-time applications.

In the remainder of the paper we review related works (\Sec{related}), and then detail our data acquisition setup and the annotation method (\Sec{dataset}). We then discuss the methods we evaluate on our dataset, as well as suggest an empirically well performing deep network architecture (\Sec{learn}).
We conclude by further detailing our experimental results (\Sec{eval}), and suggesting avenues for future works (\Sec{future}).

%%% Local Variables:
%%% mode: latex
%%% TeX-master: "../paper"
%%% End:

\section{Related works}
\label{sec:related}

We now introduce several heuristics that have been proposed for real-time hand segmentation, describe existing datasets, and overview techniques for semantic segmentation. For references on hand tracking, see~\citet{yuan2017bighand}.

\subsection{Heuristics for hand segmentation}
\label{sec:relheuristics}
The pioneering approach of \citet{oiko2011hand} leverages skin color segmentation and requires the user to wear long sleeves and to keep their face out of sight. Melax et al.~\citet{melax2013dynamics} exploited short-range depth sensors by assuming that everything within the camera field of view is to be tracked, while Oberweger et al.~\citet{oberweger2015hands} expect the hand to be the closest object to the camera. Some methods identify the ROI as the portion of the point cloud attached to the  wrist, where this can be identified either with the help of a colored wristband~\cite{tagliasacchi2015robust}, or by querying the wrist position in a full-body tracker~\cite{sharp2015accurate}.
As discussed, these heuristics fail as soon as their underlying assumptions are violated.

\subsection{Datasets for hand segmentation}
\label{sec:reldatasets}

\begin{table*}[!t]
\caption{
Existing and proposed datasets for \emph{exocentric} hand segmentation from depth imagery.
Our dataset is the only real dataset that distinguishes the two hands. Furthermore, our capture setup does not require expensive sensors as in the other two real datasets; see text for more details.
}
\begin{center}
\small
\begin{tabular}{l  lc  cc  cc }
\toprule
Dataset & Annotations & \#Frames & \#Subj & Hand & Sensor Type & Resolution
\\\midrule 
Freiburg~\cite{zimmermann2017learning} & synthetic & 43,986 & 20 & left/right & Unreal Engine & 320 $\times$ 320
\\ 
NYU \cite{tompson2014real} & automatic & 6,736 & 2 & left & Kinect v1 & 640 $\times$ 480
\\
HandNet~\cite{wetzler2015rule} & heuristic & 212,928 & 10 &  left & RealSense SR300 & 320 $\times$ 240
\\\midrule
%% \textbf{Proposed} & automatic & 265,000 & 14 &  left/right & RealSense SR300 & 640 $\times$ 480
\textbf{Proposed} & automatic & 210,000 & 13 &  left/right & RealSense SR300 & 640 $\times$ 480
\\\bottomrule
\end{tabular}
\end{center}
\label{tab:datasets}
\end{table*}
%%% Local Variables:
%%% mode: latex
%%% TeX-master: "../../paper"
%%% End:

Datasets for hand segmentation from color images were previously proposed by \citet{Buehler2008} and \citet{Bambach2015} who provided pixel-level manually annotated ground truth for respectively $\approx 500$ and $\approx 15k$ color images. Manual annotation of segmentation masks from color images is extremely labor intensive. This not only makes it very difficult to collect large-scale datasets, but the quality of annotations also depends on the skills of the individual annotator. Gathering bounding-box annotations is easier, as demonstrated by the datasets of $\approx 500$ annotated images in \citet{everingham2012pascal}, or the $\approx 15k$ images in~\citet{mittal2011hand}. However, these annotations are too coarse for applications that require accurate hand/background or hand/object segmentation.

\paragraph{Automatic segmentation}

Hand segmentation can be cast as a skin color segmentation problem~\cite{zimmermann2017learning}. However, segmenting this not only detects hands but also other skin regions, such as faces or forearms when the user is not wearing sleeves. Further, datasets of this kind~\cite{Solar2004,Kawulok2014} contain at most a few thousand manual annotations, which is magnitudes smaller than what is needed to train deep neural networks. Zimmermann et al.~\citet{zimmermann2017learning} recently proposed a dataset with $\approx 44k$ \emph{synthetic} images. However, it is notoriously difficult to accurately model skin colors in unconstrained lighting settings considering complex effects like subsurface scattering, making it challenging to develop segmentation methods that could work in the wild.
Conversely, hand segmentation from \emph{depth} images does not suffer this problem. Tompson et al.~\citet{tompson2014real} pioneered this approach and \emph{painted} each user hand with bright colors which are segmented and post-processed with the help of depth information.
However, while \cite{tompson2014real} contains $\approx 70k$ marker-annotated frames from three viewpoints, only $\approx 7k$ are provided with annotations suitable for hand segmentation. Furthermore, this dataset has been acquired with a Kinect~v1 sensor, which is now deprecated for hand tracking -- its \emph{long-range} configuration and its use of \emph{spatial structured light} results in a loss of small geometric features (e.g. fingertips) from the estimated depth map.

\paragraph{Segmentation via tracking}
Recent datasets targeting hands have mostly focused on acquiring annotated 3D marker locations for joints~\cite{yuan2017bighand}. Creating datasets via manual annotation is not only labor-intensive~\cite{sridhar2013multicam}, but placing markers within a noisy depth map often results in inaccurate labels. Assuming marker locations are correct, simple heuristics can be employed to infer a dense labeling.
Following this idea, Wetzler et al.~\citet{wetzler2015rule} first employ a complex/invasive hardware setup comprising of \emph{magnetic sensors} attached to fingertips to acquire their locations, and obtain the segmentation mask via a depth-based flood-fill.
While the dataset by~\citet{wetzler2015rule} contains $\approx 200k$ annotated exemplars, these heuristic annotations should not be considered to be ground truth for learning a high-performance segmenter; see our evaluations in \Section{eval}.

\subsection{Semantic segmentation}
\label{sec:relsegment}
Recently, neural networks have been successfully applied to the problem of semantic segmentation of a broad range of real world objects and scenes. Popular methods include fully convolutional neural networks \cite{fcnn}, which encode the input to a low-dimensional latent space, and decode via bilinear upsampling to predict the semantic segmentation.
Follow-up works perform learning at the decoder level as well, such as the well known \emph{DeconvNet}~\cite{deconvnet} and~\emph{SegNet}~\cite{segnet} architectures; see \Section{learn}. Learned encoder-decoder architectures have been shown to perform well on semantic segmentation~\cite{Zhao2017,Lin2017,Pinheiro2016,Paszke2016}, but when fast inference time is essential, random forests are an excellent alternative due to their easy parallelization~\cite{Kontschieder2011,Shotton2008}. In human pose estimation applications, \citet{Shotton2013} inferred body part labels via random forests, which was later adopted for hand localization from depth images by \citet{tompson2014real}, and color images by \citet{zimmermann2017learning}. Recently, \cite{taylor2017articulated} employed a convolutional neural network to estimate two-hand segmentation masks for hand tracking.
In multi-view setups, effective segmentation provides a strong cue for effective tracking~\cite{liu2013markerless}, and the two tasks can even be coupled into a single optimization problem~\cite{kohli2008simultaneous}.Predicted segmentation masks can be noisy and/or coarse, and post-processing is
typically employed to remove outliers by regularizing the
segmentation~\cite{Chen2015}.A recent approach by \citet{kolkin2017training} accounts for the severity of mis-labeling by a loss encoding their spatial distribution, but this method has yet to be generalized to a multi-label classification scenario like ours. Relevant to our work is also the recent R-CNN series of works, of which the instance segmentation work by \citet{he2017maskrcnn} represents the latest installment. While combining bounding box localization with dense segmentation could be effective, it is however unclear to which extent such networks could be adapted to demanding real-time applications such as hand tracking. 

%%% Local Variables:
%%% mode: latex
%%% TeX-master: "../paper"
%%% End:

%% \input{sec/3_overview}
\section{Data acquisition and automatic annotation}
\label{sec:dataset}
 
For scalable annotation with minimal human interaction, we rely on the synchronized color/depth input of an RGBD device, and a pair of skin-tight brightly colored gloves.
As shown in \Figure{teaser}(middle), this allows a (quasi) non-invasive and cost effective setup, where we can automatically determine ground-truth labels at pixel level.
As the gloves fit the user's hand tightly, minimal geometric aberration to the depth map occurs, while the consistent color of the glove can be used to extract the hand ROI via color segmentation. After an initial color calibration session, we ask the user to perform a few motions according to the \emph{protocol} described below, and record sequences of (depth,color) image pairs at a constant 48Hz rate with an Intel RealSense SR300.
We then execute a color segmentation to generate masks with a very small false positive rate; we finally quickly discard contiguous frames containing erroneous labels via manual inspection of video -- this task is \emph{significantly} simpler than manually editing individual images.
In our process we roughly drop $10\%$ of the automatically labeled images, selected conservatively to avoid any wrong label in the dataset.

\subsection{Acquisition protocol}

Similarly to~\citet{yuan2017bighand}, we attempt to maximize the coverage of the articulation space by asking each user to assume a number of example extremal poses, while capturing the natural motion during each transition. We further move the camera during the capture process to enrich the dataset with various viewpoints. Our dataset includes complex poses, such as the ones shown in \Figure{qualitative}, where fingers are overlapping with each other.

\begin{figure}[t]
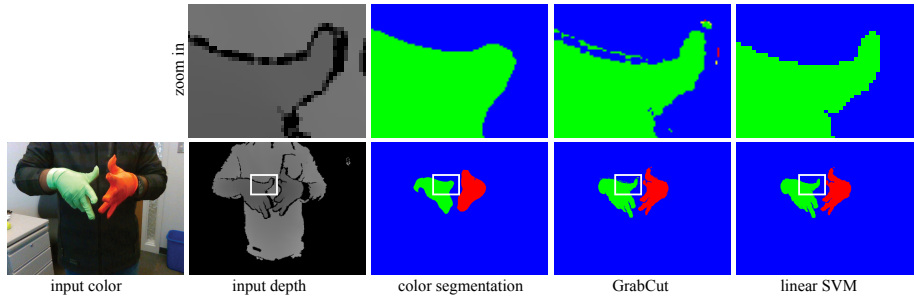

\centering
\begin{overpic} 
[width=\linewidth]
% [width=\linewidth,grid,tics=10]
%% {\currfiledir/item.pdf}
{\currfiledir/item-1.pdf}
\myfigurename{}
\end{overpic}
\caption{
Our automatic annotation pipeline. {\bf (Bottom)} input images, the output of each segmentation step. {\bf (Top)} Zoom in to highlight segmentation accuracy.
We employ the color image to create ground truth annotations for depth.
We first segment the color image via HSV thresholding, then perform GrabCut~\cite{grabcut} to obtain better segmentation. We finally train a per-image linear support vector machine (SVM) with RGB, HSV, XYZ, Lab color spaces, as well as image coordinates as input cues to further refine the annotation. Note how the segmentation becomes more accurate as each step is performed. Through this three-stage process we are able to obtain highly accurate ground-truth annotations without manual annotations. See text for details.% on each step.
}
\label{fig:segment}
\end{figure}

%%% Local Variables:
%%% mode: latex
%%% TeX-master: "../../paper"
%%% End:

\subsection{Automatic label generation}

As illustrated in \Figure{segment}, we perform color segmentation through a three-stage procedure.
The quality of the labeling is enhanced at each stage of the pipeline.

\paragraph{Initial color segmentation}
We first perform color space thresholding to obtain a rough segmentation $S_r$ and $S_l$ of the two hands, where $r$ and $l$ denote right and left hands respectively.
We will denote both $S_r$ and $S_l$ together as $S_*$. Specifically, to obtain $S_*$, we threshold on the HSV color space after smoothing the input image with a Gaussian kernel (with standard deviation 30) to remove noise. The threshold values we use for our experiments are minimum and maximum values of $[3, 160, 100]$--$[15, 255, 255]$ for left hand, and $[28, 35, 100]$--$[70, 200, 255]$ for right hand, where [H, S, V] denotes the HSV values and $H \in [0, 180]$ while $S,V \in [0, 255]$.

\paragraph{Refinement through GrabCut~\cite{grabcut}}
As the initial segmentation is coarse due to the initial Gaussian filtering, we further apply GrabCut~\cite{grabcut}, followed by a linear support vector machine (SVM) classifier~\cite{linearsvm} to get a more fine-grained segmentation map; see \Figure{segment}.
We determine the seed points for GrabCut by first finding the enclosing rectangles $R_*$ for each hand, and then using all points of $S_*$ that are inside $R_*$; 
to be robust to noise, we enlarge $R_*$ by 10 percent.
At this stage, some of the labels are still inaccurate, especially near the boundaries of the hands.

\paragraph{Refinement through linear SVM~\cite{linearsvm}}
To further enhance the labels, we exploit the high distinctiveness of the glove's color, and train a linear SVM classifier \emph{per image} with a large enough margin, and use the positives that are classified as positives during training as ground-truth labels.
Note that this classifier is simply a per-image refinement process that automatically sets-up the per-image thresholds for a simple colour-based thresholding system, based on the GrabCut results.
For robust performance we use RGB, HSV, XYZ and Lab color values as well as image coordinates as cues to linear SVM.
We also empirically set the hyper-parameter $C=900$ (margin strength).

%%% Local Variables:
%%% mode: latex
%%% TeX-master: "../paper"
%%% End:

\begin{figure}[t]
\centering
\begin{overpic} 
[width=\linewidth]
% [width=\linewidth,grid,tics=10]
{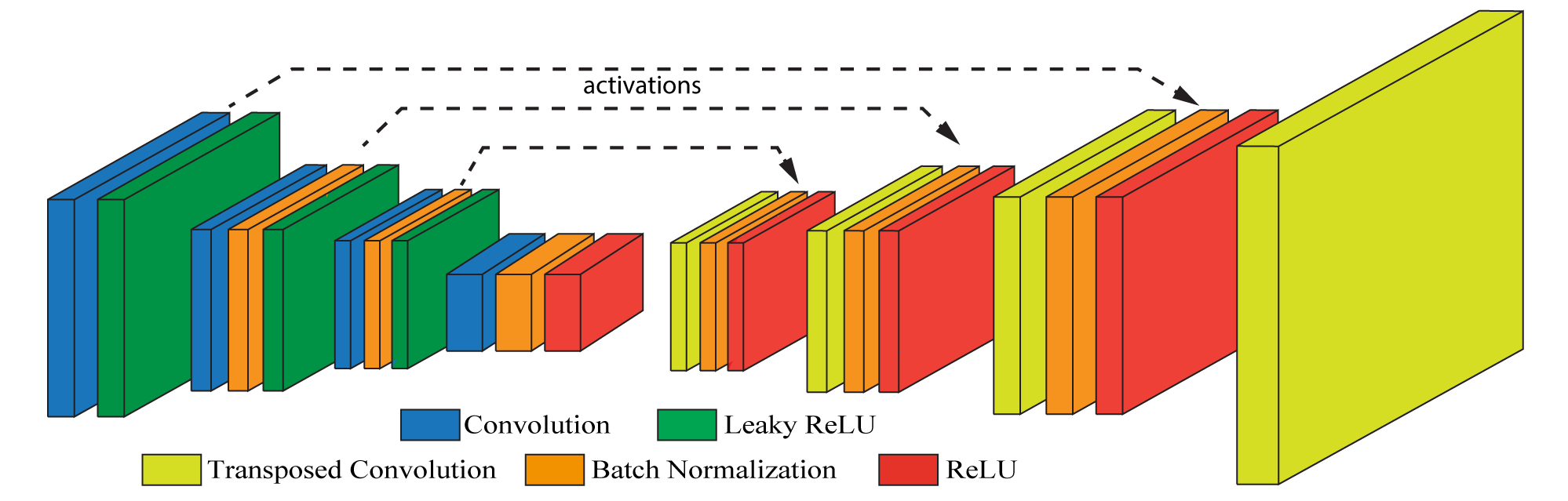}
\myfigurename{}
\end{overpic}
\caption{ %%
The architecture of our hand segmentation CNN.
Note that in our architecture we do not have any pooling/unpooling layers.
Instead, we use strided convolutions in the encoder and strided transposed convolutions in the decoder.
}
\label{fig:archs}
\end{figure}

%%% Local Variables:
%%% mode: latex
%%% TeX-master: "../../paper"
%%% End:

\section{Learning to segment hands}
\label{sec:learn}

We now detail the segmentation techniques we evaluate on our dataset in \Section{eval}.  We investigate Random Forests as a representative \emph{shallow} method (\Sec{forests}), and multiple deep architectures (\Sec{deep}).

\subsection{Random Forests}
\label{sec:forests}
Our first baseline is the \emph{shallow} learning offered by Random Forests popularized for full-body tracking by~\citet{shotton2011real}.
Tompson et al.~\citet{tompson2014real} pioneered its application to binary segmentation of one hand, while Sridhar et al.~\citet{sridhar2015fast} extended the approach to also learn more detailed part labels (e.g. palm/phalanx labels to guide articulated registration).
Analogously to \cite{shotton2011real,sridhar2015fast}, our forest consists of 3 trees each of depth 22, and uses the typical depth differential features proposed by \cite[Eq.1]{shotton2011real}. At inference time, Random Forests are highly efficient, making them suitable to applications like \emph{real-time} tracking.
However, while their optimal parameters (offset/threshold) are learned, the features themselves are fixed, and this can result in overall lower accuracy when compared to deep architectures.

\subsection{Deep convolutional segmenters}
\label{sec:deep}
 
We evaluate several recently proposed deep learning convolutional architectures, as well as propose a novel variant with enhanced forward-propagation efficiency and precision. As we have a multi-class labeling problem, we employ the soft-max cross entropy loss. In all our experiments we train our networks with ADAM optimizer with default parameters, and with appropriate learning rates between $10^{-4}$ and $10^{-6}$, depending on the architecture.
To prevent over-fitting, we apply early stopping according to the accuracy of the models on the validation set. To further improve the generalization capacity of the deep networks, we apply random data \emph{augmentation} by: randomly flipping the images horizontally as well as the left/right labels; randomly translating the depth images horizontally and vertically by 20\% proportionally to the input size; and randomly scaling the images in the range of 20\% (log scale).
We further normalize the input depth image so that the measured signals have a unit average.

\paragraph{Fully convolutional neural network (FCN)}
Long et al.~\citet{fcnn} proposed an architecture where a coarse segmentation mask is produced via a series of convolutions and max-pooling stages (\emph{encoder}), where the low-resolution image is then upsampled (\emph{decoder}) via bilinear interpolation -- the \emph{FCN32s} variant in \cite[Fig.3]{fcnn}.
As this process produces a blurry segmentation mask, a sharper mask can be obtained by combining this image with the higher-resolution activations from earlier layers in the network; the \emph{FCN16s} and \emph{FCN8s} variants. Unfortunately, the initial layers in the network only encode very localized features. Hence while this process does produce sharper results, it also introduces high-frequency mis-classifications in uncertain regions. Another problem of FCN is their difficulty in dealing with the problem of \emph{class imbalance}: in our training images, the cardinality of background pixels is significantly larger than the one of hand pixels. We overcome this problem by incorporating the class frequency in the loss~\cite{Mostajabi2015,Xu2014}, which effectively prevents the network from converging to one that trivializes the output to be always classified as background. Even with these changes, the limited accuracy achieved by this network can be understood by noting that the encoder layer is learned, while the decoder layer is not.

\paragraph{Learned encoder-decoder networks}
The popular \emph{SegNet}~\cite{segnet} and~\emph{Deconv\-Net}~\cite{deconvnet} semantic segmentation networks follow an encoder-decoder architecture. Similarly to FCNs, the encoder is realized via a sequence of convolutions and max-pooling operations. However, rather than relying on interpolation, the decoder used to generate high-resolution segmentations is also learned. Both architectures employ an \emph{unpooling} operation that inverts the \emph{max-pooling} in the encoder. Similarly to DeconvNet, SegNet upsamples the feature maps via memorized max-pooling indices in the corresponding encoder layer. Further, while unpooling in SegNet is followed by a simple series of convolutions, DeconvNet employs a series of \emph{transposed convolution} layers.
Transposed convolutions, coined ``deconvolutions'' in \cite{deconvnet}, invert the convolution process, and combined with strides are an effective way to create a feature map that is larger than the one in input.
This allows learning a segmenter with a full image, and to create a bottleneck layer that encodes the dataset manifold.
However, these architectures lack skip connections, and thus require large numbers of intermediate channels to preserve information when downsampling and upsampling. Thus, they are computationally intensive to both train and test, as we show later in the experiments, while performing worse than our architecture.

\paragraph{Proposed architecture}
We empirically found that the best performing architecture is a \emph{hybrid} encoder-decoder, see \Figure{archs}: we employ a hierarchy of transposed convolution layers (a la DeconvNet), and to improve sharpness and local detail of our predictions, without having the need for excessive amount of hidden neurons, we forward information from encoder to decoder through skip-connections (a la U-Net~\cite{Ronneberger2015}). Differently from other architectures, note how our encoders/decoders do not contain any pooling/unpooling layer. Pooling layers are useful in classification tasks as they provide invariance to local deformations, which is exactly the opposite of what we would like in our case, that is, a segmentation output that is pixel-level accurate. Nonetheless it is critical to have downsampling and upsampling for efficiency and to incorporate context in estimating the label for each pixel. In our encoder network, this is achieved by {\emph{strides}}. For the decoder network, we symmetrically employ \emph{transposed convolution} layers with strides. This enables the network to learn an appropriate upsampling filter.
The simplicity in our design results in \emph{efficient} forward propagation, while simultaneously achieving state-of-the-art \emph{accuracy};~see~\Figure{evalcrossnet} and \Table{evalcrossnet_time}

%%% Local Variables:
%%% mode: latex
%%% TeX-master: "../paper"
%%% End:

\section{Evaluation}
\label{sec:eval}
 
We quantitatively evaluate our dataset with various methods from three different angles.
In \Section{evalarchs}, we evaluate how different methods perform on our data in terms of mean Intersection over Union (mIoU), as well as their runtime both during training and testing.
In \Section{evaldatasets}, we show the generalization capabilities of several datasets, including ours.
In all our experiments, the dataset was split randomly in a 8:1:1 ratio to form train, validation, and test sets.

\paragraph{Evaluation metrics}
 
In our multi-label classification problem, each pixel can be classified as \{left, right, background\}. Within each class, we can have \emph{true-positives} (TP), \emph{false-positives} (FP) and \emph{false-negatives} (FN). Given such a categorization, we use the \emph{Intersection over Union}, defined as $\text{IoU} = {|TP|} \:/\: {\left(|TP| + |FP| + |FN|\right)}$, for quantitative evaluation.
As in \cite{zhou2017scene}, to aggregate results for multiple classes, we use the class-wise average among classes, that is, mean IoU (mIoU). This is to account for the imbalance in the number of pixels for each class.

\subsection{Segmenting with different architectures}
\label{sec:evalarchs}
\begin{figure}[t]
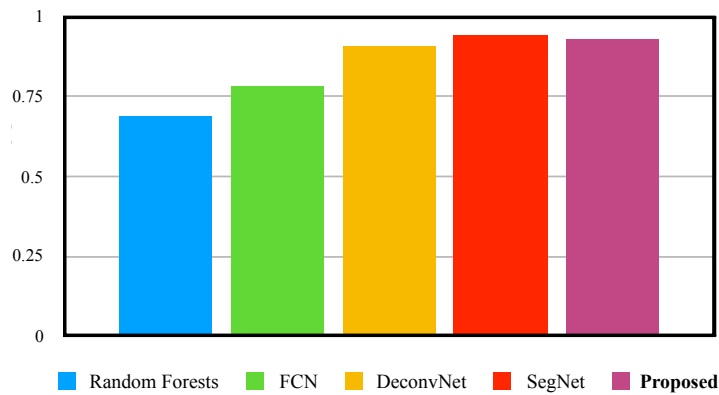

\centering

%% Figure
\begin{overpic} 
%% [width=0.8\linewidth]
[width=0.8\linewidth, trim = 24 50 24 20, clip]
% [width=\linewidth,grid,tics=10]
{\currfiledir/item-1.pdf}
\myfigurename{}
\end{overpic}

\begin{overpic} 
%% [width=0.8\linewidth]
[width=0.9\linewidth, trim = 40 20 10 240, clip]
% [width=\linewidth,grid,tics=10]
{\currfiledir/item-1.pdf}
\myfigurename{}
\end{overpic}
\caption{ Performance of different segmentation methods on our dataset in terms of mIoU (higher is better). Evaluation is performed on a two-class setup, where left and right hands are not distinguished. Otherwise, DeconvNet and SegNet fail to learn. However, the proposed network is able to achieve state of the art in any case.}

\label{fig:evalcrossnet}
\end{figure}

%% \label{tab:evalcrossnet}
%% \end{table*}
%%% Local Variables:
%%% mode: latex
%%% TeX-master: "../../paper"
%%% End:

\begin{table*}[t]
\caption{
  Runtime of each segmentation method. Ours is the fastest to train and test amongst compared deep architectures.
}
\setlength{\tabcolsep}{6pt}
\begin{center}
\normalsize
\begin{tabular}{l c c c c c}
  \toprule
  & Random Forests & FCN & DeconvNet & SegNet & {\bf Proposed} \\
  \midrule
  Train time &  3h & 149h & 57h & 83h & 29h \\
  Test time &  1ms & 41ms & 16ms & 30ms & 5ms \\
  \bottomrule
\end{tabular}
\end{center}
\label{tab:evalcrossnet_time}
\end{table*}
%%% Local Variables:
%%% mode: latex
%%% TeX-master: "../../paper"
%%% End:

In \Figure{evalcrossnet}, we compare the different learning approaches in terms of accuracy, and their runtime in \Table{evalcrossnet_time}. For the runtime experiments, all deep networks were run on a single NVIDIA Geforce GTX 1080 Ti graphics card. In these experiments, we did not distinguish between left and right hands, as DeconvNet and SegNet completely failed due to the class imbalance between left hand, right hand, and background labels.
Although Random Forests are clearly the fastest to train and to infer on, they perform poorly when compared to deep networks.
Due to its simple upsampling scheme, FCN(32s) performs the worst among the evaluated networks.
Thanks to its learned decoder network, DeconvNet and SegNet obtain much better results. However, their architectures are too computationally complex, resulting in a runtime that is not suitable for real-time tracking applications when considering that segmentation is typically a pre-processing step for a sophisticated vision pipeline. Our proposed architecture not only on par with the best performing method in terms of accuracy, but it is also fast to forward-propagate, running at $\approx 200$fps. Furthermore, as noted previously, when we train networks to distinguish between left and right hands, our architecture, HandSeg, is the only network among the best three that gives any usable result (mIoU 0.877, as shown in \Table{crossdataset_dual}).

\subsection{Cross-dataset evaluation}
\label{sec:evaldatasets}
\begin{figure}[t]
\centering

%% Figure
\begin{overpic} 
[width=0.65\linewidth]
% [width=\linewidth,grid,tics=10]
{\currfiledir/item-1.pdf}
\myfigurename{}
\end{overpic}
%% Figure
\begin{overpic} 
[width=0.26\linewidth, trim = 0 -5 0 0, clip]
% [width=\linewidth,grid,tics=10]
{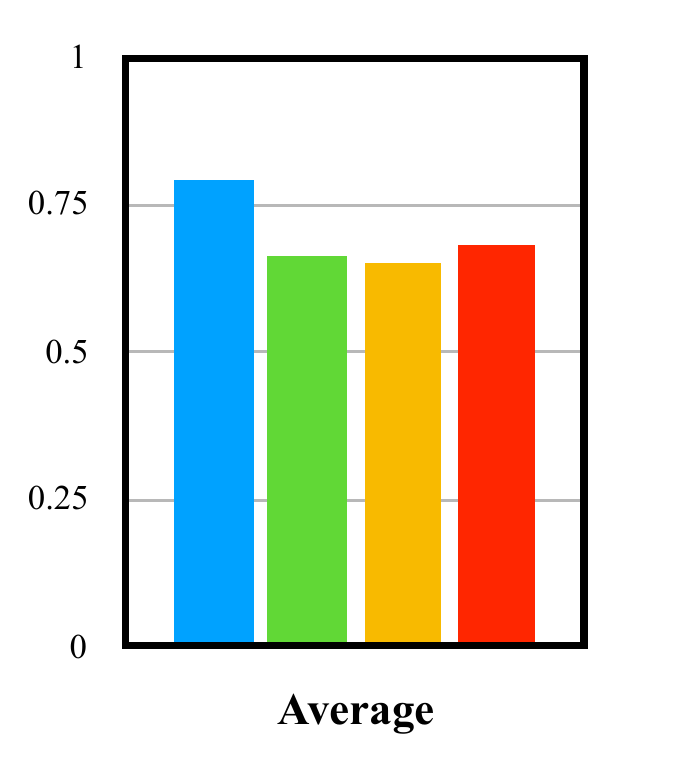}
\end{overpic}

\vspace{0.5em}
\begin{overpic} 
[width=0.8\linewidth]
% [width=\linewidth,grid,tics=10]
{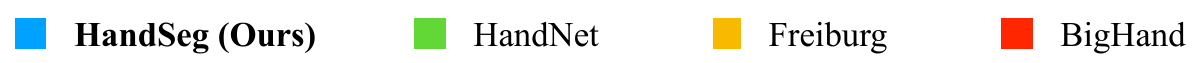}
\end{overpic}
\caption{
Generalization performance across datasets for the \textbf{two-class setup}, in terms of mIoU. The dataset used for training is color-coded by the legend at the bottom, and the results are grouped by each test set. The washed-out colors denote the case when trained and tested on the same dataset. On the right, we show the average performance of segmenters trained on each dataset, when tested on other datasets excluding the one used for training. Note that the segmenter trained on our dataset, \emph{HandSeg}, generalizes best on average and on Freiburg, and is on par with the best generalizing dataset on HandNet.
%\TODO{Change both legends -- {\bf HandSeg} $\rightarrow$ {\bf HandSeg} (Ours)}
% 
%\TODO{@Abhi, Update the figures so that the number ranges from 0 to 1. mIoU should not be a percent value.}
% 
}
\label{fig:crossdataset}
\end{figure}

We test our baseline network by training/testing on all possible combinations of the datasets in \Table{datasets} (with the exception of NYU which is captured using a deprecated sensor).
We also include BigHands~\cite{yuan2017bighand} as a dataset for training, which is a hand pose dataset composed of more than a million images.
However, as this dataset is originally intended for hand pose estimation, it does not have per pixel labels.
We therefore apply GrabCut~\cite{grabcut} with the hands' joint locations as seed points to obtain a rough segmentation label.
As these labels are not perfect, this dataset cannot be used for testing. As not all datasets distinguish left and right hands, we perform evaluations for the two-class (hands vs. background) as well as the three-class (left vs. right vs. background) scenario.
We summarize the results in \Figure{crossdataset} and in \Table{crossdataset_dual}, respectively.

%% \begin{table}[t]
%% \begin{center}
%% \begin{tabular}{lcccc}
%%     \toprule
%%     \diagbox[width=.6in]{Train}{Test} & \textbf{HandSeg} & HandNet & Freiburg \\
%%     \midrule
%%     %NYU & \textbf{91.6} & \emph{69.8} & 44.8 & 67.8 \\
%%     \textbf{HandSeg} & \textbf{93.01} & 83.77 & 75.74 \\
%%     HandNet & 67.38 & \textbf{98.54} & 65.99 \\
%%     Freiburg & 61.28 & 69.30  & \textbf{94.39} \\
%%     Bighand & 69.84 & 86.11 & 66.00 \\
%%     \bottomrule 
%% \end{tabular}
%% \end{center}
%% \label{tab:crossdataset}
%% \caption{Generalization across datasets (mIoU).}
%% \end{table}
%%% Local Variables:
%%% mode: latex
%%% TeX-master: "../../paper"
%%% End:

As shown in \Figure{crossdataset}, when left and right hands are not distinguished, the segmenter trained with our dataset generalizes better (in average) than when trained with other datasets.
Furthermore, as shown in the results on each testing dataset, the segmenter trained on our dataset performs either the best or comparably to the best method, while simultaneously generalizing to unseen datasets.

%% \begingroup
\begin{table}[t]
\caption{
Generalization performance across datasets for the \textbf{three-class setup}, in terms of mIoU. For BigHands, we use data augmentation to generate both left and right hand labels. Segmenter trained on our dataset, \emph{HandSeg}, performs best in terms of generalization.
}
\begin{center}
\setlength{\tabcolsep}{6pt}
\begin{tabular}{lcccc}
    \toprule
    \diagbox[width=.9in]{Test}{Train} & \textbf{HandSeg} (Ours) & Freiburg & BigHands\\
    \midrule
  \textbf{HandSeg} (Ours) & 0.877 & 0.437 & 0.492 \\
  Freiburg & 0.574 & 0.870 & 0.408 \\
    \bottomrule 
\end{tabular}
\end{center}
\label{tab:crossdataset_dual}
\end{table}
%% \endgroup
%%% Local Variables:
%%% mode: latex
%%% TeX-master: "../../paper"
%%% End:

%% 
In \Table{crossdataset_dual}, we show the case when the two hands are distinguished. This three-class case is harder than the two-class case above, as the segmenter now has to distinguish between left and right hands. As shown, the segmenter trained with our dataset generalizes better to Freiburg, than the one trained with Freiburg on ours. Considering that the test performance on both datasets is similar, this shows the better generalization capability of our dataset.
Furthermore, as the BigHands dataset only features a single hand, data augmentation needs to be applied for the three-class setup, which is the result shown in \Table{crossdataset_dual}. The poor numbers clearly demonstrate the need of a hand segmentation dataset that distinguishes left/right hands.

\subsection{Qualitative evaluation}
\begin{figure*}[!t]
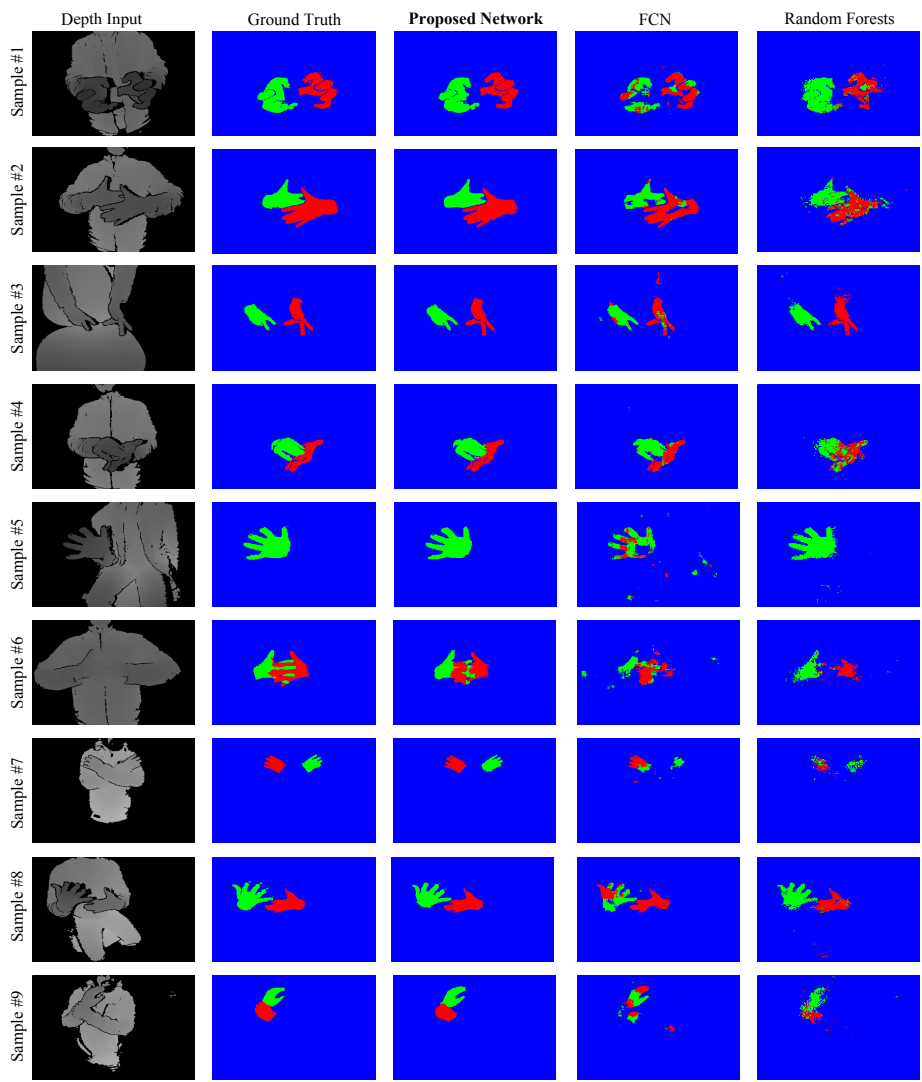

\centering
\begin{overpic} 
%% [width=\linewidth]
[width=\linewidth, trim = 5 70 10 0, clip] %remove last 1 examples
%% [width=\linewidth, trim = 0 70 0 0, clip] %remove last 1 examples
%% [width=\linewidth, trim = 0 130 0 0, clip] %remove last 2 examples
%% [width=\linewidth, trim = 0 200 0 0, clip] %remove last 3 examples
%% [width=\linewidth, trim = 0 260 0 0, clip] %remove last 4 examples
% [width=\linewidth,grid,tics=10]
{\currfiledir/item-1.pdf}
\myfigurename{}
\end{overpic}
\vspace{0.01cm}
\caption{
Qualitative examples. We illustrate a few examples of hand segmentation performance on our dataset for the proposed network, FCN, and Random Forests. We exclude SegNet and DeconvNet here as they converge to estimating all pixel as the background for this setup. Note how the proposed network gives accurate segmentation for diverse poses, including when the hands are interacting as shown in Sample~\#4. Sample~\#6 shows a failure case of our network when there's extreme interaction between the two hands. Still, our architecture performs better than the compared ones, giving relatively accurate segmentation.
  %\TODO{ABI, please provide 4/5 more rows of data? So we can fill the 14th page better}
  }
\label{fig:qualitative}
\end{figure*}

%%% Local Variables:
%%% mode: latex
%%% TeX-master: "../../paper"
%%% End:

In \Figure{qualitative}, we provide qualitative segmentation results on our novel dataset.
Here, we show results of the proposed architecture, FCN, and the Random Forest. We excluded SegNet and DeconvNet, as for the three-class experiments, these two network architectures failed to deliver any meaningful results on our dataset and converged to a trivial solution, that is, all pixels considered as background.
Note how the proposed architecture shows the best performance.

\begin{figure}[t]
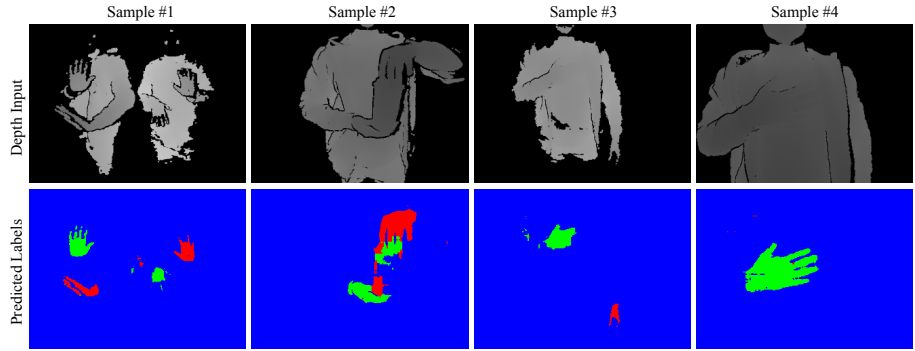

\centering
\begin{overpic} 
[width=\linewidth]
% [width=\linewidth,grid,tics=10]
%% {\currfiledir/item.pdf}
{\currfiledir/item-1.pdf}
\myfigurename{}
\end{overpic}
\caption{
A selection of segmentation failure cases.
Due to the challenging nature of these examples, our segmenter does not return perfect results. Note that in Sample~\#1, our network is able to segment all four hands, although it was never trained with more than a single person in the field of view. In Sample~\#2, our network show error on the cup, as the network never saw hands interacting with objects.
%% 
%\TODO{@Abhi, replace cad* files with actual depth instead of color on depth}
% 
}
\label{fig:failures}
\end{figure}

\Figure{failures} shows challenging frames where our network does not deliver perfect results.
Sample \#1 illustrates how the network can still segment the hands of multiple persons, although it was trained on frames containing a single individual. This reveals the generalization capabilities of our network, which did not only learn to segment \textit{one/two} regions, but also learned a latent \emph{shape-space} for human hands. Sample~\#2 shows a person holding a cup, while Sample~\#3 and Sample~\#4 have the hand lying flat on the body. These scenarios are difficult, as the network has never seen a hand interacting with objects.
Although not perfect, the network successfully segments the hands in Samples \#3 and \#4, but fails on the cup for Sample~\#2.
Accuracy could be improved by accounting for the additional information in the color channel, or by learning the appearance of the object via training examples.

%%% Local Variables:
%%% mode: latex
%%% TeX-master: "../paper"
%%% End:

\section{Conclusions and future works}
\label{sec:future}
We have proposed an automatic annotation method for easily creating hand segmentation datasets with an RGBD camera, and have introduced a new high-quality dataset for hand segmentation that is significantly larger than what is currently available. Our annotation method requires minimal human interaction, and is highly cost effective. With the proposed method, we have created a dataset that contains high-accuracy dense pixel annotations, large pose variations, and many different subjects. Our results show that the new dataset, HandSeg, allows training of segmenters that are more general than the ones trained with existing datasets.

Our analysis has also revealed poor generalization characteristics for currently available methods. With the Microsoft Kinect~v1 sensor being retired from production, this creates an immediate problem as the only high-quality (albeit small) dataset for the task at hand~\cite{tompson2014real} becomes unusable. Conversely, our data is acquired on Intel RealSense SR300 sensors, one of the most commonly employed sensors available. Beyond these immediate needs, it would also be interesting to see whether simultaneously training on multiple datasets could generate architectures that are apt to transfer learning. While eventually the use of (very large) synthetic datasets like~\cite{zimmermann2017learning} could be very effective for training, the proposed HandSeg dataset will remain valuable for validation/testing. 

We also propose a segmentation network that is faster than existing baselines, and provides superior mIoU accuracy. While these results are encouraging, our dataset opens new frontiers for investigation, such as the effectiveness of spatially-aware losses~\cite{kolkin2017training}, the use of efficient quantized networks~\cite{bdnn}, or its use for weak-supervision of discriminative hand tracking~\cite{Neverova2017}.

%%% Local Variables:
%%% mode: latex
%%% TeX-master: "../../paper"
%%% End:

%%% Local Variables:
%%% mode: latex
%%% TeX-master: "../paper"
%%% End:

\newpage
% {
%     \small
%     \bibliographystyle{ieee}
%     \bibliography{paper}
% }

%{
 %    \small
 %   \bibliographystyle{abbrvnat} %< \citet works!!
 %    \bibliography{paper,markus}
% }
\bibliographystyle{splncs}
\bibliography{paper,markus}
% \MO{Attention: ACCV allows only 2 pages of references this year!}
% \AT{thanks for the catch!!}

%%% Local Variables:
%%% mode: latex
%%% TeX-master: "../paper"
%%% End:

% \clearpage
\end{document}